\documentclass[11pt]{article}

\usepackage[hyperref]{naacl2021}
\usepackage{url}
\usepackage{times}
\usepackage{latexsym}
\usepackage{booktabs}
\usepackage{subcaption}
\usepackage[T1]{fontenc}
\usepackage[utf8]{inputenc}
\usepackage{microtype}
\usepackage{multirow}
\usepackage{tabularx}
\usepackage{capt-of}
\usepackage{xspace}
\newcommand{\F}{F$_1$\xspace}
\usepackage{graphicx}
\usepackage{enumitem}
\setlist[description]{leftmargin=0cm,labelindent=0cm,itemsep=5pt,parsep=0pt}
\newenvironment{example}{\begin{quote}\sffamily\small}{\end{quote}}

\title{Claim Detection in Biomedical Twitter Posts}

\author{Amelie W\"uhrl \and Roman Klinger \\
  Institut f{\"u}r Maschinelle Sprachverarbeitung, University of Stuttgart, Germany \\
  \texttt{\{amelie.wuehrl,roman.klinger\}@ims.uni-stuttgart.de}\\
}

\begin{document}
\maketitle
\begin{abstract}
  Social media contains unfiltered and unique information, which is
  potentially of great value, but, in the case of misinformation, can
  also do great harm. With regards to biomedical topics, false
  information can be particularly dangerous. Methods of automatic
  fact-checking and fake news detection address this problem, but have
  not been applied to the biomedical domain in social media yet. We
  aim to fill this research gap and annotate a corpus of 1200 tweets
  for implicit and explicit biomedical claims (the latter also with
  span annotations for the claim phrase). With this corpus, which we
  sample to be related to COVID-19, measles, cystic fibrosis, and
  depression, we develop baseline models which detect tweets that
  contain a claim automatically. Our analyses reveal that biomedical
  tweets are densely populated with claims (45\,\% in a corpus sampled
  to contain 1200 tweets focused on the domains mentioned above).
  Baseline classification experiments with embedding-based classifiers
and BERT-based transfer learning demonstrate that the detection
is challenging, however, shows acceptable
performance for the identification of explicit expressions of claims. Implicit claim tweets
are more challenging to detect.
\end{abstract}

\section{Introduction}

Social media platforms like Twitter contain vast amounts of valuable
and novel information, and biomedical aspects are no exception
\citep{Correia2020}. Doctors share insights from their everyday life,
patients report on their experiences with particular medical
conditions and drugs, or they discuss and hypothesize about the
potential value of a treatment for a particular disease. This
information can be of great value -- governmental administrations or
pharmaceutical companies can for instance learn about unknown side
effects or potentially beneficial off-label use of medications.

\begin{figure}
    \centering
    \includegraphics[width=0.7\linewidth]{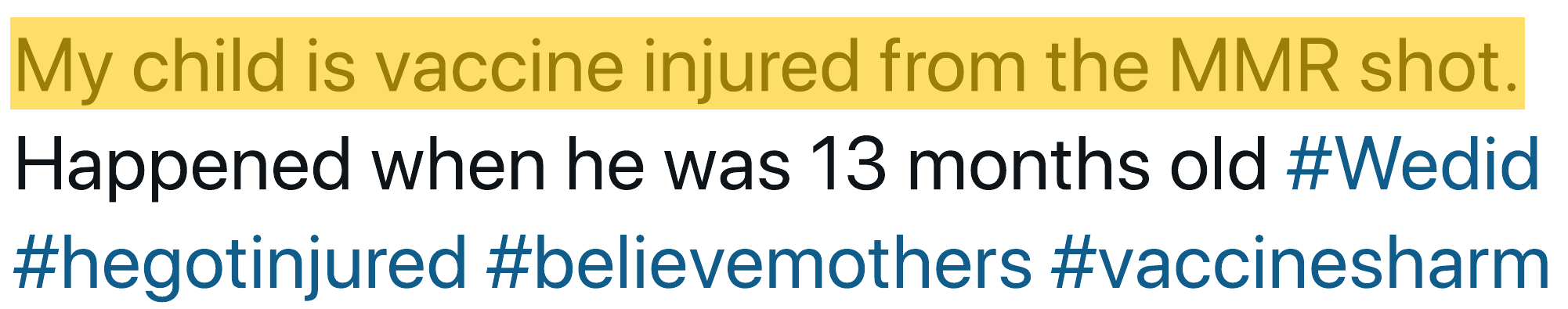}
    \caption{Tweet with a biomedical claim (highlighted).}
    \label{fig:example-claim-tweet}
\end{figure}

At the same time, unproven claims or even intentionally spread
misinformation might also do great harm. Therefore, contextualizing a
social media message and investigating if a statement is debated or
can actually be proven with a reference to a reliable resource is
important. The task of detecting such claims is essential in argument
mining and a prerequisite in further analysis for tasks like
fact-checking or hypotheses generation. We show an example of a tweet
with a claim in Figure~\ref{fig:example-claim-tweet}.

Claims are widely considered the conclusive and therefore central part
of an argument \citep{Lippi2015,Stab2017}, consequently making it the
most valuable information to extract. Argument mining and claim
detection has been explored for texts like legal documents, Wikipedia
articles, essays \citep[i.a.]{Moens2007,Levy2014,Stab2017}, social
media and web content \citep[i.a.]{Goudas2014,Habernal2017,
  Bosc2016,Dusmanu2017}. It has also been applied to scientific
biomedical publications \citep[i.a.]{Achakulvisut2019,Mayer2020}, but
biomedical arguments as they occur on social media, and particularly
Twitter, have not been analyzed yet.

With this paper, we fill this gap and explore claim detection for
tweets discussing biomedical topics, particularly tweets about
COVID-19, the measles, cystic fibrosis, and depression, to allow for
drawing conclusions across different
fields.

Our contributions to a better understanding of biomedical claims made
on Twitter are, (1), to publish the first biomedical Twitter corpus
manually labeled with claims (distinguished in explicit and implicit,
and with span annotations for explicit claim phrases), and (2),
baseline experiments to detect (implicit and explicit) claim tweets in
a classification setting. Further, (3), we find in a cross-corpus
study that a generalization across domains is challenging and that
biomedical tweets pose a particularly difficult environment for claim
detection.

\section{Related Work}
Detecting biomedical claims on Twitter is a task rooted in both the
argument mining field as well as the area of biomedical text mining.

\subsection{Argumentation Mining}

Argumentation mining covers a variety of different domains, text, and
discourse types. This includes online content, for instance Wikipedia
\cite{Levy2014,Roitman2016,Lippi2015}, but also more
interaction-driven platforms, like fora. As an example,
\citet{Habernal2017} extract argument structures from blogs and forum
posts, including comments. Apart from that, Twitter is generally a
popular text source \citep{Bosc2016,Dusmanu2017}.  Argument mining is
also applied to professionally generated content, for instance news
\cite{Goudas2014,Sardianos2015} and legal or political documents
\cite{Moens2007,Palau2009,Mochales2011,Florou2013}.  Another domain of
interest are persuasive essays, which we also use in a cross-domain
study in this paper \cite{Lippi2015,Stab2017,Eger2017}.

Existing approaches differ with regards to which tasks in the broader
argument mining pipeline they address. While some focus on the
detection of arguments
\cite{Moens2007,Florou2013,Levy2014,Bosc2016,Dusmanu2017,Habernal2017},
others analyze the relational aspects between argument components
\cite{Mochales2011,Stab2017,Eger2017}.

While most approaches cater to a specific domain or text genre,
\citet{Stab2018} argue that domain-focused, specialized systems do not
generalize to broader applications such as argument search in
texts. In line with that, \citet{Daxenberger2017} present a
comparative study on cross-domain claim detection. They observe that
diverse training data leads to a more robust model performance in
unknown domains.

\subsection{Claim Detection}
\label{sec:rel-work-general-claim-detection}
Claim detection is a central task in argumentation mining. It can be
framed as a classification (Does a document/sentence contain
a claim?) or as sequence labeling (Which tokens make up the
claim?). The setting as classification has been explored, inter alia, as a
retrieval task of online comments made by public stakeholders on
pending governmental regulations \citep{Kwon2007}, for sentence
detection in essays, \citep{Lippi2015}, and for Wikipedia
\cite{Roitman2016,Levy2017}. The setting as a sequence labeling task
has been tackled on Wikipedia \cite{Levy2014}, on Twitter, and on
news articles \cite{Goudas2014,Sardianos2015}.

One common characteristic in most work on automatic claim detection is
the focus on relatively formal text. Social media, like
tweets, can be considered a more challenging text type, which despite
this aspect, received considerable attention, also beyond
classification or token sequence labeling.  \newcite{Bosc2016} detect
relations between arguments, \citet{Dusmanu2017} identify factual or
opinionated tweets, and \citet{Addawood2016} further classify the type
of premise which accompanies the claim. \citet{Ouertatani2020} combine
aspects of sentiment detection, opinion, and argument mining in a
pipeline to analyze argumentative tweets more comprehensively.
\citet{Ma2018} specifically focus on the claim detection task in
tweets, and present an approach to retrieve Twitter posts that contain
argumentative claims about debatable political topics.

To the best of our knowledge, detecting biomedical claims in tweets
has not been approached yet. Biomedical argument mining, also for
other text types, is generally still limited. The work by \citet{Shi2019}
is one of the few exceptions that target this challenge and propose a
pipeline to extract health-related claims from headlines of
health-themed news articles. The majority of other argument mining
approaches for the biomedical domain focus on research literature
\citep{Blake2010,Alamri2015,Alamri2015b,Achakulvisut2019,Mayer2020}.
\begin{table*}
    \centering\footnotesize
    \begin{tabularx}{\linewidth}{XXXX}
    \toprule
    \multicolumn{4}{c}{Query category}\\
    \cmidrule{1-4}
    \multicolumn{1}{c}{Disease Names} &\multicolumn{1}{c}{Topical Hashtags} &\multicolumn{1}{c}{Combinations } &\multicolumn{1}{c}{Drugs}     \\
    \cmidrule(r){1-1}\cmidrule(lr){2-2}\cmidrule(lr){3-3}\cmidrule(l){4-4}
    
    COVID-19, \#COVID-19 
    & \#socialdistancing,\par \#chinesevirus
    & COVID-19 AND cured, COVID-19 AND vaccines
    & Hydroxychloroquine,\par Kaletra, Remdesivir\\
    
    measles, \#measles 
    &\#vaccineswork,\par \#dontvaccinate
    & measles AND vaccine, measles AND therapize
    & M-M-R II, Priorix,\par ProQuad \\
    
    cystic fibrosis,\par \#cysticfibrosis
    & \#livesavingdrugs4cf, \#orkambinow
    & cystic fibrosis AND treated, cystic fibrosis AND heal
    & Orkambi, Trikafta,\par Tezacaftor\\
    
    depression, \#depression
    & \#depressionisreal,\par \#notjustsad
    &depression AND cure,\par depression AND treatment
    & Alprazolam, Buspirone,\par Xanax \\
    \bottomrule
    \end{tabularx}
    \caption{Examples of the four categories of search terms used to
      retrieve tweets about COVID-19, the measles, cystic fibrosis,
      and depression via the Twitter API.}
    \label{tab:search-terms-Twitter-api}
\end{table*} 

\subsection{Biomedical Text Mining}
Biomedical natural language processing (BioNLP) is a field in
computational linguistics which also receives substantial attention
from the bioinformatics community. One focus is on the automatic
extraction of information from life science articles, including entity
recognition, e.g., of diseases, drug names, protein and gene names
\citep[i.a.]{Habibi2017,Giorgi2018,Lee2019} or relations between those
\citep[i.a.]{Lamurias2019,Sousa2021,Lin2019}.

Biomedical text mining methods have also been applied to social media texts
and web content \citep[i.a.]{Wegrzyn2011,Yang2016,Sullivan2016}. One focus is on
the analysis of Twitter with regards to pharmacovigilance. Other topics include the
extraction of adverse drug reactions \citep{Nikfarjam2015,Cocos2017},
monitoring public health \citep{Paul2012,Choudhury2013,Sarker2016},
and detecting personal health mentions \citep{Yin2015,Karisani2018}.

A small number of studies looked into the comparison of biomedical
information in social media and scientific text: \citet{Thorne2018}
analyze quantitatively how disease names are referred to across these
domains. \citet{Seiffe2020} analyze laypersons' medical vocabulary.

\section{Corpus Creation and Analysis}
As the basis for our study, we collect a novel Twitter corpus in
which we annotate which tweets contain biomedical claims, and (for all
explicit claims) which tokens correspond to that claim.

\subsection{Data Selection \& Acquisition}

The data for the corpus was collected in June/July 2020 using
Twitter's
API\footnote{\url{https://developer.twitter.com/en/docs/twitter-api}}
which offers a keyword-based retrieval for tweets. Table
\ref{tab:search-terms-Twitter-api} provides a sample of the search
terms we used.\footnote{The full list of search terms (1771 queries in
  total) is available in the supplementary material.} For each of the
medical topics, we sample English tweets from keywords and phrases
from four different query categories. This includes (1) the name of
the disease as well as the respective hashtag for each topic, e.g.,
\textit{depression} and \textit{\#depression}, (2) topical hashtags
like \textit{ \#vaccineswork}, (3) combinations of the disease name
with words like \textit{cure, treatment} or \textit{therapy} as well
as their respective verb forms, and (4) a list of medications,
products, and product brand names from the pharmaceutical database
DrugBank\footnote{\url{https://go.drugbank.com/}. At the time of
  creating the search term list, COVID-19 was not included in
  DrugBank. Instead, medications which were under investigation at the
  time of compiling this list as outlined on the WHO website were
  included for Sars-CoV-2 in this category:
  \url{https://www.who.int/emergencies/diseases/novel-coronavirus-2019/global-research-on-novel-coronavirus-2019-ncov/solidarity-clinical-trial-for-covid-19-treatments}.}.

When querying the tweets, we exclude retweets by using the API's
`-filter:retweets' option. From overall 902,524 collected tweets, we
filter out those with URLs since those are likely to be advertisements
\citep{Cocos2017,Ma2018}, and further remove duplicates based on the
tweet IDs. From the resulting collection of 127,540 messages we draw a
sample of 75 randomly selected tweets per topic (four biomedical
topics) and search term category (four categories per topic).  The
final corpus to be annotated consists of 1200 tweets about four
medical issues and their treatments: measles, depression, cystic
fibrosis, and COVID-19.

\subsection{Annotation}

\subsubsection{Conceptual Definition}
While there are different schemes and models of argumentative
structure varying in complexity as well as in their conceptualization
of claims, the claim element is widely considered the core component
of an argument \citep{Daxenberger2017}. \citet{Aharoni2014} suggest a
framework in which an argument consists of two main components: a
claim and premises. We follow \citet{Stab2017} and define the claim as
the argumentative component in which the speaker or writer expresses
the central, controversial conclusion of their argument. This claim is
presented as if it were true even though objectively it can be true or
false \citep{Mochales2009}. The premise which is considered the second
part of an argument includes all elements that are used either to
substantiate or disprove the claim. Arguments can contain multiple
premises to justify the claim. (Refer to Section
\ref{sec:qual_analysis} for examples and a detailed analysis of
argumentative tweets in the dataset.)

For our corpus, we focus on the claim element and assign all tweets a
binary label that indicates whether the document contains a
claim.  Claims can be either explicitly voiced or the claim property
can be inferred from the text in cases in which they are expressed
implicitly \citep{Habernal2017}. We therefore annotate explicitness or
implicitness if a tweet is labeled as containing a claim.  For
explicit cases the claim sequence is additionally marked on the token
level.  For implicit cases, the claim which can be inferred from the
implicit utterance is stated alongside the implicitness annotation.

\subsubsection{Guideline Development}
We define a preliminary set of annotation guidelines based on previous
work
\cite{Mochales2009,Aharoni2014,Bosc2016,Daxenberger2017,Stab2017}.  To
adapt those to our domain and topic, we go through four iterations of
refinements. In each iteration, 20 tweets receive annotations by two
annotators. Both annotators are female and aged 25--30. Annotator A1
has a background in linguistics and computational linguistics. A2 has
a background in mathematics, computer science, and computational
linguistics. The results are discussed based on the calculation of
Cohen's $\kappa$ \citep{Cohen1960}.

After Iteration 1, we did not make any substantial changes, but
reinforced a common understanding of the existing guidelines in a
joint discussion.
After Iteration 2, we clarified the guidelines by adding
the notion of an argumentative intention as a pre\-requisite for a
claim: a claim is only to be annotated if the author actually appears
to be intentionally argumentative as opposed to just sharing an
opinion \citep{Snaijder2016, Habernal2017}.
This is illustrated in the following example, which is not to be
annotated as a claim, given this additional constraint:
\begin{example}
  This popped up on my memories from two years ago, on Instagram, and
  honestly I’m so much healthier now it’s quite unbelievable. A stone
  heavier, on week 11 of no IVs (back then it was every 9 weeks), and
  it’s all thanks to \#Trikafta and determination. I am stronger than
  I think.
\end{example}

We further clarified the guidelines with regards to the claim being
the conclusive element in a Twitter document. This change encouraged
the annotators to reflect specifically if the conclusive, main claim
is conveyed explicitly or implicitly.

After Iteration 3, we did not introduce any changes, but went through
an additional iteration to further establish the understanding of the
annotation tasks.

\begin{table}
  \centering\small
  \setlength{\tabcolsep}{10pt}
\begin{tabular}{lccc}
  \toprule
  & \multicolumn{3}{c}{Cohen's $\kappa$}\\
  \cmidrule{2-4}
  & C/N & E/I/N & Span \\
  \cmidrule(r){2-2}\cmidrule(lr){3-3}\cmidrule(l){4-4}
  Iteration 1       & .31 & .43 & .32\\
  Iteration 2       & .34 & .24 & .12\\
  Iteration 3       & .61 & .42 & .42\\
  Iteration 4       & .60 & .68 & .41 \\
  \cmidrule(r){2-2}\cmidrule(lr){3-3}\cmidrule(l){4-4}
  Final corpus      & .56 & .48 & .38\\
  \bottomrule
\end{tabular}
\caption{Inter-annotator agreement during development of the
  annotation guidelines and for the final corpus. C/N:
  Claim/non-claim, E/I/N: Explicit/Implicit/Non-claim, Span: Token-level
  annotation of the explicit claim expression.}
\label{tab:corpus-iaa}
\end{table}

Table~\ref{tab:corpus-iaa} shows the results of the agreement of the
annotators in each iteration as well as the final $\kappa$-score for
the corpus. We observe that the agreement substantially increased from
Iteration 1 to 4. However, we also observe that obtaining a
substantial agreement for the span annotation remains the most
challenging task.

\subsubsection{Annotation Procedure}

The corpus annotation was carried out by the same annotators that
conducted the preliminary annotations.  A1 labeled 1000 tweets while
A2 annotated 300 instances. From these both sets, 100 tweets were
provided to both annotators, to track agreement (which remained
stable, see Table~\ref{tab:corpus-iaa}). Annotating 100 tweets took
approx.\ 3.3 hours. Overall, we observe that the agreement is
generally moderate. Separating claim-tweets from non-claim tweets
shows an acceptable $\kappa$=.56. Including the decision of
explicitness/implicitness leads to $\kappa$=.48. The span-based
annotation has limited agreement, with $\kappa$=.38 (which is why we
do not consider this task further in this paper).  These numbers are
roughly in line with previous work.  \citet{Achakulvisut2019} report
an average $\kappa$=0.63 for labeling claims in biomedical research
papers.  According to \citet{Habernal2017}, explicit, intentional
argumentation is easier to annotate than texts which are less
explicit.

Our corpus is available with detailed annotation guidelines at
\url{http://www.ims.uni-stuttgart.de/data/bioclaim}.

\subsection{Corpus Statistics}
Table \ref{tab:corpus-class-distributions} presents corpus
statistics. Out of the 1200 documents in the corpus, 537 instances
(44.75\,\%) contain a claim and 663 (55.25 \%) do not. From all claim
instances, 370 tweets are explicit (68 \%). The claims are not
equally distributed across topics (not shown in table): 61\,\% of
measle-related tweets contain a claim, 49\,\% of those related to
COVID-19, 40\,\% of cystic fibrosis tweets, and 29\,\% for depression.

\begin{table}
  \centering\small
  \setlength{\tabcolsep}{9pt}
\begin{tabular}{lccc}
\toprule
Class & \# Instances & \% & Length \\
\cmidrule(r){1-1}\cmidrule(lr){2-2}\cmidrule(lr){3-3}\cmidrule(l){4-4}
  non-claim    & 663 & 55.25 & 30.56 \\
  claim (I+E)  & 537 & 44.75 & 39.88 \\
  expl.\ claim & 370 & 30.83 & 39.89 \\
  \ \ claim phrase &     &       & 17.59 \\
  impl.\ claim & 167 & 13.92 & 39.88 \\
\cmidrule(r){1-1}\cmidrule(lr){2-2}\cmidrule(lr){3-3}\cmidrule(l){4-4}
total & 1200 & 100 \% & 34.73 \\
\bottomrule
\end{tabular}
\caption{Distribution of the annotated classes and average instance lengths
  (in tokens).}
\label{tab:corpus-class-distributions}
\end{table}

\begin{table}
  \centering\small
  \setlength{\tabcolsep}{6pt}
    \begin{tabular}{lrrrrrrrrrr}
      \toprule
         & \multicolumn{2}{c}{incompl.} & \multicolumn{2}{c}{blended} & \multicolumn{2}{c}{anecdotal} & \multicolumn{2}{c}{impl.} \\
          \cmidrule(l){2-3} \cmidrule(l){4-5} \cmidrule(l){6-7} \cmidrule(l){8-9} \cmidrule(l){10-11} 
         M
         & 8  & .16
         & 14 & .28
         & 9  & .18
         & 14 & .28
         \\
         C 
         & 17 & .34  
         & 15 & .30  
         & 8  & .16  
         & 14 & .28   
         \\
         CF 
         & 12 & .24  
         & 10 & .20  
         & 26 & .52  
         & 18 & .36  
         \\
         D 
         & 16 & .32  
         & 9 & .18  
         & 23 & .46  
         & 11 & .22 
         \\
         \cmidrule(l){1-1} \cmidrule(l){2-3} \cmidrule(l){4-5} \cmidrule(l){6-7} \cmidrule(l){8-9} \cmidrule(l){10-11} 
         total 
         & 53 & .27 
         & 48 & .24  
         & 66 & .33  
         & 57 & .29 
         \\
         \bottomrule
    \end{tabular}
    \caption{Manual analysis of a subsample of 50 tweets/topic. Each
      column shows raw counts and percentage/topic.}
    \label{tab:num-claims-per-topic}
\end{table}

The longest tweet in the corpus consists of 110 tokens\footnote{The
  tweet includes 50 @-mentions followed by a measles-related claim:
  ``Oh yay! I can do this too, since you're going to ignore the
  thousands of children who died in outbreaks last year from
  measles... Show me a proven death of a child from vaccines in the
  last decade. That's the time reference, now? So let's see a death
  certificate that says it, thx''}, while the two shortest consist only
of two tokens\footnote{``Xanax damage'' and ``Holy fuck''.}. On average,
a claim tweet has a length of $\approx$40 tokens. Both claim tweet types, explicit and
implicit, have similar lengths (39.89 and 39.88 tokens
respectively). In contrast to that, the average non-claim tweet is
shorter, consisting of about 30 tokens. Roughly half of an explicit
claim corresponds to the claim phrase.

We generally see that there is a connection between the length of a
tweet and its class membership. Out of all tweets with up to 40
tokens, 453 instances are non-claims, while 243 contain a claim. For
the instances that consist of 41 and more tokens, only 210 are
non-claim tweets, whereas 294 are labeled as claims. The majority of
the shorter tweets ($\leq$ 40 tokens) tend to be non-claim instances,
while mid-range to longer tweets ($\geq$ 40 tokens) tend to be members
of the claim class.

\begin{table}
  \centering\small
  \begin{tabularx}{1.0\linewidth}{lX}
    \toprule
    id & Instance \\
    \cmidrule(r){1-1}\cmidrule(l){2-2}
    1 & \textit{The French have had great success \#hydroxycloroquine.} \\
    2 & Death is around 1/1000 in measles normally, same for
        encephalopathy, hospitalisation around 1/5. \textit{With all
        the attendant costs, the vaccine saves money, not makes it.}\\
    3 & Latest: Kimberly isn’t worried at all. \textit{She takes \#Hydroxychloroquine and feels awesome the next day.} Just think, it’s more dangerous to drive a car than to catch corona \\
    4 & Lol exactly. It’s not toxic to your body idk where he pulled this information out of.           \textit{Acid literally cured my depression/anxiety I had for 5 years in just 5 months (3 trips).} It literally reconnects parts of your brain that haven’t had that connection in a long time. \\
    5 & Hopefully! The MMR toxin loaded vaccine I received many years ago seemed to work very well.     More please! \\
    6 & Wow! Someone tell people with Cystic fibrosis and Huntington’s
        that they can cure their genetics through Mormonism! \\
    \bottomrule
  \end{tabularx}
  \caption{Examples of explicit and implicit claim tweets from the corpus. Explicit claims are in italics.}
  \label{tab:examples}
\end{table}

\subsection{Qualitative Analysis}
\label{sec:qual_analysis}

To obtain a better understanding of the corpus, we perform a qualitative
analysis on a subsample of 50 claim-instances/topic. We manually
analyze four claim properties: the tweet exhibits an incomplete
argument structure, different argument components blend into each
other, the text shows anecdotal evidence, and it describes the claim
implicitly. Refer to Table \ref{tab:num-claims-per-topic}
for an overview of the results.

In line with \citet{Snaijder2016}, we find that
argument structures are often incomplete, e.g., instances only contain a
stand-alone claim without any premise. This characteristic is most
prevalent in the COVID-19-related tweets In Table~\ref{tab:examples},
Ex.~1 is missing a premising element, Ex.\ 2 presents premise
and claim.

Argument components (claim, premise) are not very clear cut and often
blend together. Consequently they can be difficult to differentiate,
for instance when authors use claim-like elements as a premise. This
characteristic is again, most prevalent for COVID-19. In Ex.~3 in
Table~\ref{tab:examples}, the last sentence reads like a claim,
especially when looked at in isolation, yet it is in fact used by the
author to explain their claim.

Premise elements which substantiate and give reason for the claim
\citep{Bosc2016b} traditionally include references to studies or
mentions of expert testimony, but occasionally also anecdotal evidence
or concrete examples \cite{Aharoni2014}. We find the latter to be
very common for our data set. This property is most frequent for
cystic fibrosis and depression. Ex.~4 showcases how a personal
experience is used to build an argument.

Implicitness in the form of irony, sarcasm or rhetoric questions are
common features for these types of claims on Twitter. We observe
claims related to cystic fibrosis are most often (in our sample)
implicit. Ex.\ 5 and 6 show instances that use sarcasm or irony.
The fact that implicitness is such a common feature in our dataset is
in line with the observation that implicitness is a characteristic
device not only in spoken language and everyday, informal
argumentation \citep{Lumer1990}, but also in user-generated
web content in general \citep{Habernal2017}.

\section{Methods}
\label{sec:methods}

In the following sections we describe the conceptual design of our
experiments and introduce the models that we use to accomplish the
claim detection task.

\subsection{Classification Tasks}
\label{sec:tasks}
We model the task in a set of different model configurations.
\begin{description}
\item [Binary.] A trained classifier distinguishes between claim
and non-claim.
\item [Multiclass.] A trained classifier distinguishes between
exlicit claim, implicit claim, and non-claim.
\item [Multiclass Pipeline.] A first classifier learns to
discriminate between claims and non-claims (as in Binary). Each tweet
that is classified as claim is further separated into implicit or
explicit with another binary classifier. The secondary classifier is
trained on gold data (not on predictions of the first model in the
pipeline).
\end{description}


\begin{table*}
  \centering\small
    \begin{tabular}{lllcccccccccccc}
    \toprule
    \multicolumn{3}{l}{} & \multicolumn{3}{c}{NB} & \multicolumn{3}{c}{LG} & \multicolumn{3}{c}{LSTM} & \multicolumn{3}{c}{BERT}\\
    \cmidrule(ll){4-6}
    \cmidrule(ll){7-9}
    \cmidrule(ll){10-12}
    \cmidrule(ll){13-15}
     Eval. & Task & Class
     & \multicolumn{1}{c}{P} & \multicolumn{1}{c}{R} & \multicolumn{1}{c}{\F} 
     & \multicolumn{1}{c}{P} & \multicolumn{1}{c}{R} & \multicolumn{1}{c}{\F}
     & \multicolumn{1}{c}{P} & \multicolumn{1}{c}{R} & \multicolumn{1}{c}{\F}
     & \multicolumn{1}{c}{P} & \multicolumn{1}{c}{R} & \multicolumn{1}{c}{\F}\\
      \cmidrule(r){1-1}
      \cmidrule(lr){2-2}
      \cmidrule(lr){3-3}
      \cmidrule(lr){4-6}
      \cmidrule(lr){7-9}
      \cmidrule(lr){10-12}
      \cmidrule(l){13-15}
    
     \multirow{4}{*}{\rotatebox[origin=tr]{90}{~binary~}} 
     & \multirow{2}{*}{binary} & claim 
     & .67 & .65 & .66
     & .66 & .74 &\textbf{.70}
     & .68 & .48 & .57
     & .66 & .72 & .69 \\
     
     & & n-claim
     &.75& .77&  \textbf{.76}
     &.79&.72& \textbf{.76}
     &.69 & .84 &.75 
     & .78 & .72 & .75 \\[5pt]
     
     & \multirow{2}{*}{multiclass} & claim 
     &.66 & .65 & \textbf{.66}
     &.73 & .53 & .61
     &.75 & .35 & .48
     & .81 & .49 & .61 \\
     
     & & n-claim
     &.74 & .76 &.75
     &.71 & .85 & .78
     &.66 & .91 & .76
     & .71 & .91 & \textbf{.80}\\
    
      \cmidrule(r){1-3}
      \cmidrule(lr){4-6}
      \cmidrule(lr){7-9}
      \cmidrule(lr){10-12}
      \cmidrule(l){13-15}

    \multirow{6}{*}{\rotatebox[origin=tr]{90}{~multi-class~}} & \multirow{3}{*}{multiclass} & expl 
    &.55 & .45 & .50
    &.63 & .39 & .48
    &.59 & .27 & .37
    &.62 & .45 & \textbf{.52} \\
    
    &  & impl 
    & .31 &.44 &\textbf{.36}
    & .33& .35 & .34
    & .18 & .09 & .12
    & .29 & .09 & .13 \\
    
    & & n-claim
     &.74 & .76 & .75
     &.71 & .85 & .78
     &.66 & .91 & .76
     & .71 & .91 & \textbf{.80}\\[5pt]
    
    & \multirow{3}{*}{pipeline} & expl 
    & .56 & .45 & .50
    & .52 & .55 & .53
    & .50 & .37 & .43
    & .54 & .65 & \textbf{.59} \\
    
    &  & impl 
    & .31 &.44 &\textbf{.36}
    & .28 &.35 &.31
    & .07 &.04 & .05
    & .26 & .22 & .24 \\
    
    & & n-claim
     &.75 & .77&\textbf{ .76}
     &.79 & .72 & \textbf{.76}
     &.69 & .84 & .75
     & .78 & .72 & .75\\
    \bottomrule
    \end{tabular}
  \caption{Results for the claim detection experiments, separated into
    binary and multi-class evaluation. The best \F scores for each
    evaluation setting and class are printed in bold face.}
  \label{tab:experiments-results}
\end{table*}

\subsection{Model Architecture}
\label{sec:arch}
For each of the classification tasks (binary/multiclass, steps in the
pipeline), we use a set of standard text classification methods which
we compare. The first three models (NB, LG, BiLSTM) use 50-dimensional
FastText \citep{Bojanowski2017} embeddings trained on the Common Crawl
corpus (600 billion tokens) as
input\footnote{\url{https://fasttext.cc/docs/en/english-vectors.html}}.

\begin{description}
\item[NB.] We use a (Gaussian) naive
  Bayes with an average vector of the token embeddings as input.
\item[LG.] We use a logistic regression classifier with the same
  features as in NB. 
\item[BiLSTM.] As a classifier which can consider contextual
  information and makes use of pretrained embeddings, we use a
  bidirectional long short-term memory network \citep{Hochreiter1997}
  with 75 LSTM units followed by the output layer (sigmoid for binary
  classification, softmax for multiclass).
\item[BERT.]  We use the pretrained BERT \citep{Devlin-etal-2019} base
  model\footnote{\url{https://huggingface.co/bert-base-uncased}} and
  fine-tune it using the claim tweet corpus.
\end{description}

\section{Experiments}
\subsection{Claim Detection}
\label{sec:experiment-claim-detection}

With the first experiment we explore how reliably we can detect claim
tweets in our corpus and how well the two different claim types
(\textit{explicit} vs.\ \textit{implicit claim tweets}) can be
distinguished. We use each model mentioned in Section~\ref{sec:arch}
in each setting described in Section~\ref{sec:tasks}.  We evaluate
each classifier in a binary or (where applicable) in a multi-class
setting, to understand if splitting the claim category into its
subcomponents improves the claim prediction overall.

\subsubsection{Experimental Setting}
From our corpus of 1200 tweets we use 800 instances for training, 200
as validation data to optimize hyperparameters and 200 as test data.
We tokenize the documents and substitute all @-mentions by
``@username''. For the LG models, we use an l2 regularization. For the
LSTM models, the hyper-parameters learning rate, dropout, number of
epochs, and batch size were determined by a randomized search over a
parameter grid and also use l2 regularization. For training, we use
Adam \citep{Kingma2015}. For the BERT models, we experiment with
combinations of the recommended fine-tuning hyper-parameters from
\citet{Devlin-etal-2019} (batch size, learning rate, epochs), and use
those with the best performance on the validation data. An overview of
all hyper-parameters is provided in Table~\ref{tab:hyper-params} in
the Appendix.  For the Bi\-LSTM, we use the Keras API
\citep{Chollet2015} for TensorFlow \citep{Abadi2015}. For the BERT
model, we use Simple Transformers \citep{simpletransformers2019} and
its wrapper for the Hugging Face transformers library
\citep{wolf-etal-2020}. Further, we oversample the minority class of
implicit claims to achieve a balanced training set (the test set
remains with the original distribution). To ensure comparability, we
oversample in both the binary and the multi-class setting. For
parameters that we do not explicitly mention, we use default values.

\subsubsection{Results}

Table~\ref{tab:experiments-results} reports the results for the
conducted experiments. The top half lists the results for the binary
claim detection setting. The bottom half of the table presents the
results for the multi-class claim classification.

For the binary evaluation setting, we observe that casting the problem
as a ternary prediction task is not beneficial -- the best \F score is
obtained with the binary LG classifier (.70\,\F for the class claim in
contrast to .61\,\F for the ternary LG). The BERT and NB approaches are
slightly worse (1 pp and 4pp less for binary, respectively), while the
LSTM shows substantially lower performance (13pp less).

In the ternary/multi-class evaluation, the scores are overall
lower. The LSTM shows the lowest performance. The best result
is obtained in the pipeline setting, in which separate classifiers can
focus on distinguishing claim/non-claim and explicit/implicit -- we
see .59\,\F for the explicit claim class. Implicit claim detection is
substantially more challenging across all classification approaches.

We attribute the fact that the more complex models (LSTM, BERT) do not
outperform the linear models across the board to the comparably small
size of the dataset. This appears especially true for implicit claims
in the multi-class setting. Here, those models struggle the most to
predict implicit claims, indicating that they were not able to learn
sufficiently from the training instances.

\subsubsection{Error Analysis}

From a manual introspection of the best performing model in the binary
setting, we conclude that it is difficult to detect general
patterns. We show two cases of false positives and two cases of false
negatives in Table~\ref{tab:errors}. The false positive instances show
that the model struggles with cases that rely on
judging the argumentative intention. Both Ex.~1 and 2 contain
potential claims about depression and therapy, but they have not been
annotated as such, because the authors' intention is 
motivational rather than argumentative.  In addition, it appears that the
model struggles to detect implicit claims that are expressed using
irony (Ex.~3) or a rhetorical question (Ex.\ 4).

\begin{table}
  \centering\small
  \setlength{\tabcolsep}{2pt}
  \renewcommand{\arraystretch}{1.4}
  \begin{tabularx}{1.0\linewidth}{lllX}
    \toprule
    id & G & P & Text \\
    \cmidrule(r){1-1}\cmidrule(l){2-2}\cmidrule(l){3-3}\cmidrule(l){4-4}
    1 & n & c & \#DepressionIsReal \#MentalHealthAwareness \#mentalhealth ruins lives. \#depression destroys people. Be there when someone needs you. It could change a life. It may even save one.\\
    2 & n & c & The reason I stepped away from twitch and gaming with friends is because iv been slowly healing from a super abusive relationship. Going to therapy and hearing you have ptsd isnt easy. But look how far iv come, lost some depression weight and found some confidence:)plz stay safe \\
    3 & c & n & Not sure who knows more about \#COVID19, my sister or \#DrFauci.  My money is on Stephanie.\\
    4 & c & n &  How does giving the entire world a \#COVID19 \#vaccine compare to letting everyone actually get \#covid? What would you prefer? I'm on team @username \#WHO \#CDC \#math \#VaccinesWork \#Science \\
    \bottomrule
  \end{tabularx}
  \caption{Examples of incorrect predictions by the LG model in the
    binary setting (G:Gold, P:Predictions; n: no claim; c: claim).}
  \label{tab:errors}
\end{table}

\subsection{Cross-domain Experiment}
We see that the models show acceptable performance in a binary
classification setting. In the following, we analyze if this
observation holds across domains or if
information from another out-of-domain corpus can help.

As the binary LG model achieved the best results during the previous
experiment, we use this classifier for the cross-domain experiments. We
work with paragraphs of persuasive essays \citep{Stab2017} as a
comparative corpus. The motivation to use this resource is that while
they are a distinctly different text type and usually linguistically
much more formal than tweets, they are also opinionated
documents.\footnote{An essay is defined as \textit{``a short piece of
    writing on a particular subject, often expressing personal
    views''}
  (\url{https://dictionary.cambridge.org/dictionary/english/essay}).}
We use the resulting essay model for making an in-domain as well as a
cross-domain prediction and vice versa for the Twitter model. We
further experiment with combining the training portions of both
datasets and evaluate its performance for both target domains.

\subsubsection{Experimental Setting}

The comparative corpus contains persuasive essays with annotated
argument structure \citep{Stab2017}. \citet{Eger2017} used this corpus
subsequently and provide the data in CONLL-format, split into
paragraphs, and predivided into train, development and test
set.\footnote{\url{https://github.com/UKPLab/acl2017-neural_end2end_am/tree/master/data/conll/Paragraph_Level}}
We use their version of the corpus.  The annotations for the essay
corpus distinguish between major claims and claims. However, since
there is no such hierarchical differentiation in the Twitter
annotations, we consider both types as equivalent. We choose to use
paragraphs instead of whole essays as the individual input documents
for the classification and assign a claim label to every paragraph
that contains a claim.  This leaves us with 1587 essay paragraphs as
training data, and 199 and 449 paragraphs respectively for validation
and testing.

We follow the same setup as for the binary setting in the first experiment.

\subsubsection{Results}
In Table \ref{tab:experiments-results-comparative}, we summarize the
results of the cross-domain experiments with the persuasive essay
corpus.
\begin{table}
  \centering\small
    \begin{tabular}{llccc}
    \toprule
    Train  & Test & P & R & \F \\
    \cmidrule(r){1-1}\cmidrule(lr){2-2}\cmidrule(lr){3-3}\cmidrule(lr){4-4}\cmidrule(l){5-5}
    Twitter & Twitter & .66 & .74 &  .70 \\
    Essay & Twitter &.51 & .69 & .59 \\
    Twitter+Essay & Twitter & .58 & .75 & .66\\
    Essay & Essay &.96 & 1.0 & .98\\
    Twitter & Essay & .94 & .74 & .83\\
    Twitter+Essay & Essay &.95 & 1.0 & .97\\
    \bottomrule
    \end{tabular}
  \caption{Results of cross-domain experiments using the binary LG model on the Twitter and the essay corpus. We report precision, recall and \F for the claim tweet class.}
  \label{tab:experiments-results-comparative}
\end{table}
We see that the essay model is successful for classifying claim
documents (.98\,\F) in the in-domain experiment. Compared to the
in-domain setting for tweets all evaluation scores measure
substantially higher.

When we compare the two cross-domain experiments, we observe that the
performance measures decrease in both settings when we use the
out-of-domain model to make predictions (11pp in \F for tweets, 15pp
for essays).  Combining the training portions of both data sets does
not lead to an improvement over in-domain experiments. This shows the
challenge of building domain-generic models that perform well across
different data sets.

\section{Discussion and Future Work}
In this paper, we presented the first data set for biomedical claim
detection in social media. In our first experiment, we showed that we
can achieve an acceptable performance to detect claims when the
distinction between explicit or implicit claims is not considered. In
the cross-domain experiment, we see that text formality, which is one
of the main distinguishing feature between the two corpora, might be
an important factor that influences the level of difficulty in
accomplishing the claim detection task.

Our hypothesis in this work was that biomedical information on Twitter
exhibits a challenging setting for claim detection. Both our
experiments indicate that this is true.  Future work needs to
investigate what might be reasons for that. We hypothesize that our
Twitter dataset contains particular aspects that are specific to the
medical domain, but it might also be that other latent variables
lead to confounders (e.g., the time span that has been used for
crawling). It is important to better understand these properties.

We suggest future work on claim detection models optimize those to
work well across domains. To enable such research, this paper
contributed a novel resource. This resource could further be improved.
One way of addressing the moderate agreement between the annotators
could be to include annotators with medical expertise to see if this
ultimately facilitates claim annotation.  Additionally, a detailed
introspection of the topics covered in the tweets for each disease
would be interesting for future work since this might shed some light
on which topical categories of claims are particularly difficult to
label.

The COVID-19 pandemic has sparked recent research with regards to
detecting misinformation and fact-checking claims (e.g.,
\citet{Hossain_2020} or \citet{Wadden_2020}). Exploring how a
claim-detection-based fact-checking approach rooted in argument mining
compares to other approaches is up to future research.

\subsection*{Acknowledgments}
This research has been conducted as part of the FIBISS
project\footnote{Automatic Fact Checking for Biomedical Information in
  Social Media and Scientific Literature,
  \url{https://www.ims.uni-stuttgart.de/en/research/projects/fibiss/}}
which is funded by the German Research Council (DFG, project number:
KL 2869/5-1). We thank Laura Ana Maria Oberl\"ander for her support
and the anonymous reviewers for their valuable comments.

\appendix

\begin{table}

\begin{subtable}{\linewidth}
\centering
   \begin{tabularx}{\linewidth}{Xll}
    \toprule
    Parameter & LSTM2 & LSTM3 \\
    \cmidrule(r){1-1}
    \cmidrule(lr){2-2}
    \cmidrule(l){3-3}
    Embedding & fastText & fastText \\
    Emb.\ Dim. & 50 & 50 \\
    \# LSTM units & 75 &75\\
    Training epochs & 60 & 70 \\
    Training batch size &10 &30\\
    Loss function & Binary CE& Categ. CE\\
    Optimizer & Adam & Adam\\
    Learning rate& 1e-3 & 1e-3\\
    L2 regularization &1e-3 & 1e-3\\
    dropout &0.5& 0.6\\
    \bottomrule
    \end{tabularx}
   \caption{Overview of architectural choices and hyper-parameter
    settings for the binary (LSTM2) and multi-class (LSTM3) LSTM-based
    models used in our experiments. }\label{tab:hyper-params-LSTM}
\end{subtable}

\bigskip
\begin{subtable}{\linewidth}
\centering
   \begin{tabularx}{\linewidth}{Xll}
    \toprule
    Parameter & BERT2 & BERT3 \\
    \cmidrule(r){1-1}
    \cmidrule(lr){2-2}
    \cmidrule(l){3-3}
    Training epochs & 4 & 4 \\
    Training batch size &16 &16\\
    Learning rate&2e-5 & 3e-5\\
    \bottomrule
    \end{tabularx}
   \caption{Overview of fine-tuning hyper-parameters for the binary (BERT2) and multi-class (BERT3) models used in our experiments.}\label{tab:hyper-params-BERT}
\end{subtable}

\caption{Overview of model hyper-parameters.} \label{tab:hyper-params}
\end{table}

\end{document}